\documentclass[10pt,twocolumn,letterpaper]{article}

\usepackage{cvpr}
\usepackage{times}
\usepackage{epsfig}
\usepackage{graphicx}
\usepackage{amsmath}
\usepackage{amssymb}
\usepackage{pgfplotstable}
\usepackage{pifont}   
\usepackage{nicefrac}
\usepackage{multirow}

\usepackage[pagebackref=true,breaklinks=true,letterpaper=true,colorlinks,bookmarks=false]{hyperref}

\cvprfinalcopy 

\def\cvprPaperID{2779} 

\ifcvprfinal\fi

\usepackage{pgfplots}
\usepackage{pgfplotstable}
\usepackage{pgf,tikz}
\usetikzlibrary{arrows,shapes,decorations.markings}
\pgfplotsset{compat=1.9}

\newcommand{\extfig}[2]{\tikzsetnextfilename{fig/extern/#1}{#2}}
\newcommand{\extdata}[1]{\input{#1}}

\newcommand{\leg}[1]{\addlegendentry{#1}}

\begin{document}

\title{Mining on Manifolds: Metric Learning without Labels}

\author{
Ahmet Iscen$^1$ \ \ \ \ Giorgos Tolias$^1$\ \ \ \ Yannis Avrithis$^2$\ \ \ \ Ond{\v r}ej Chum$^{1}$\\
{\fontsize{11}{13}\selectfont$^1$VRG, FEE, CTU in Prague\ \ \ \ \ \ $^2$Inria Rennes}\\
}

\maketitle

\newcommand{\real}{\mathbb{R}}
\newcommand{\realnn}{{\mathbb{R}^{+}_{0}}}
\newcommand{\nat}{\mathbb{N}}
\newcommand{\natzero}{{\mathbb{N}_{0}}}

\newcommand{\mypar}[1]{\noindent \textbf{#1}}
\newcommand{\head}[1]{{\smallskip\noindent\bf #1}}

\def\l2{\ensuremath{\ell_2}\xspace}
\newcommand{\loss}{\ensuremath{\mathcal{L}}\xspace}

\newcommand{\se}{\ensuremath{s_{e}}\xspace}
\newcommand{\sm}{\ensuremath{s_{m}}\xspace}

\newcommand{\nne}[1]{\ensuremath{\text{NN}_{#1}^{e}}\xspace}
\newcommand{\nnd}[1]{\ensuremath{\text{NN}_{#1}^{m}}\xspace}

\def\iff{\Longleftrightarrow}

\renewcommand{\paragraph}[1]{{\medskip \noindent \bf #1}}
\newcommand{\pari}[1]{{\medskip \noindent \it #1}}
\newcommand{\equ}[1]{Equation~(\ref{#1})\xspace}

\newcommand{\alert}[1]{{\color{red}{#1}}}
\newcommand{\ahmet}[1]{{\color{blue}{#1}}}

\renewcommand{\b}[1]{\textbf{#1}}
\newcommand{\w}[1]{\color{blue}{#1}}

\def\sssp{\hspace{1pt}}
\def\ssp{\hspace{3pt}}
\def\msp{\hspace{5pt}}
\def\bsp{\hspace{8pt}}
\def\nsp{\hspace{-2pt}}

\newcommand{\cmark}{\ding{51}}%
\newcommand{\xmark}{\ding{55}}%

\newcommand{\diag}{\operatorname{diag}}
\newcommand{\vone}{\mathbf{1}}

\newcommand{\cA}{\mathcal{A}}
\newcommand{\cB}{\mathcal{B}}
\newcommand{\cC}{\mathcal{C}}
\newcommand{\cD}{\mathcal{D}}
\newcommand{\cE}{\mathcal{E}}
\newcommand{\cF}{\mathcal{F}}
\newcommand{\cG}{\mathcal{G}}
\newcommand{\cH}{\mathcal{H}}
\newcommand{\cI}{\mathcal{I}}
\newcommand{\cJ}{\mathcal{J}}
\newcommand{\cK}{\mathcal{K}}
\newcommand{\cL}{\mathcal{L}}
\newcommand{\cM}{\mathcal{M}}
\newcommand{\cN}{\mathcal{N}}
\newcommand{\cO}{\mathcal{O}}
\newcommand{\cP}{\mathcal{P}}
\newcommand{\cQ}{\mathcal{Q}}
\newcommand{\cR}{\mathcal{R}}
\newcommand{\cS}{\mathcal{S}}
\newcommand{\cT}{\mathcal{T}}
\newcommand{\cU}{\mathcal{U}}
\newcommand{\cV}{\mathcal{V}}
\newcommand{\cW}{\mathcal{W}}
\newcommand{\cX}{\mathcal{X}}
\newcommand{\cY}{\mathcal{Y}}
\newcommand{\cZ}{\mathcal{Z}}

\newcommand{\va}{\mathbf{a}}
\newcommand{\vb}{\mathbf{b}}
\newcommand{\vc}{\mathbf{c}}
\newcommand{\vd}{\mathbf{d}}
\newcommand{\ve}{\mathbf{e}}
\newcommand{\vf}{\mathbf{f}}
\newcommand{\vg}{\mathbf{g}}
\newcommand{\vh}{\mathbf{h}}
\newcommand{\vi}{\mathbf{i}}
\newcommand{\vj}{\mathbf{j}}
\newcommand{\vk}{\mathbf{k}}
\newcommand{\vl}{\mathbf{l}}
\newcommand{\vm}{\mathbf{m}}
\newcommand{\vn}{\mathbf{n}}
\newcommand{\vo}{\mathbf{o}}
\newcommand{\vp}{\mathbf{p}}
\newcommand{\vq}{\mathbf{q}}
\newcommand{\vr}{\mathbf{r}}
\newcommand{\vs}{\mathbf{s}}
\newcommand{\vt}{\mathbf{t}}
\newcommand{\vu}{\mathbf{u}}
\newcommand{\vv}{\mathbf{v}}
\newcommand{\vw}{\mathbf{w}}
\newcommand{\vx}{\mathbf{x}}
\newcommand{\vy}{\mathbf{y}}
\newcommand{\vz}{\mathbf{z}}

\newcommand{\vpi}{\boldsymbol{\pi}}

\begin{abstract}
In this work we present a novel unsupervised framework for hard training example mining. The only input to the method is a collection of images relevant to the target application and a meaningful initial representation, provided \eg by pre-trained CNN.
Positive examples are distant points on a single manifold, while negative examples are nearby points on different manifolds. Both types of examples are revealed by disagreements between Euclidean and manifold similarities.
The discovered examples can be used in training with any discriminative loss.

The method is applied to unsupervised fine-tuning of pre-trained networks
for fine-grained classification and particular object retrieval. Our models are on par or are outperforming prior models that are fully or partially supervised.

\vspace{-10pt}
\end{abstract}

\section{Introduction}
\label{sec:intro}

The success of deep neural networks on large-scale problems has been first demonstrated on the task of supervised classification~\cite{KSH12}. It was shown that embeddings, typically provided by the convolutional layers of a network, are applicable beyond classification tasks. These tasks include particular object retrieval~\cite{GARL16}, local descriptor learning~\cite{HLJ+15}, ranking~\cite{WSL+14}, and nearest-neighbor regression~\cite{BSSO16}.
A common practice is to start with a pre-trained network~\cite{SZ14,SLJS+15,HZRS16} and apply metric learning~\cite{CHL05,WSL+14,HKC+17} to fine-tune the network for a particular task.

To improve over the initial network, novel training samples are sought for which the initial network performs poorly. Such training samples are used to re-train the network using loss functions alternative to cross-entropy (\eg contrastive~\cite{CHL05}, triplet~\cite{WSL+14} or batch-level~\cite{OXJS16}).
The approaches to obtain relevant training data range from further human supervision~\cite{BSCL14,OXJS16} to instance clustering~\cite{RPDB15,MeHD17,BaSO17}, exploiting the temporal dimension in video~\cite{WG15,1708.02901}, predicting the spatial layout of image patches~\cite{Doersch_2017_ICCV}, or using existing computer vision pipelines to match unstructured image collections pairwise~\cite{GARL16,RTC16}.

\begin{figure}
\small
\begin{center}
\includegraphics[height=1.7cm]{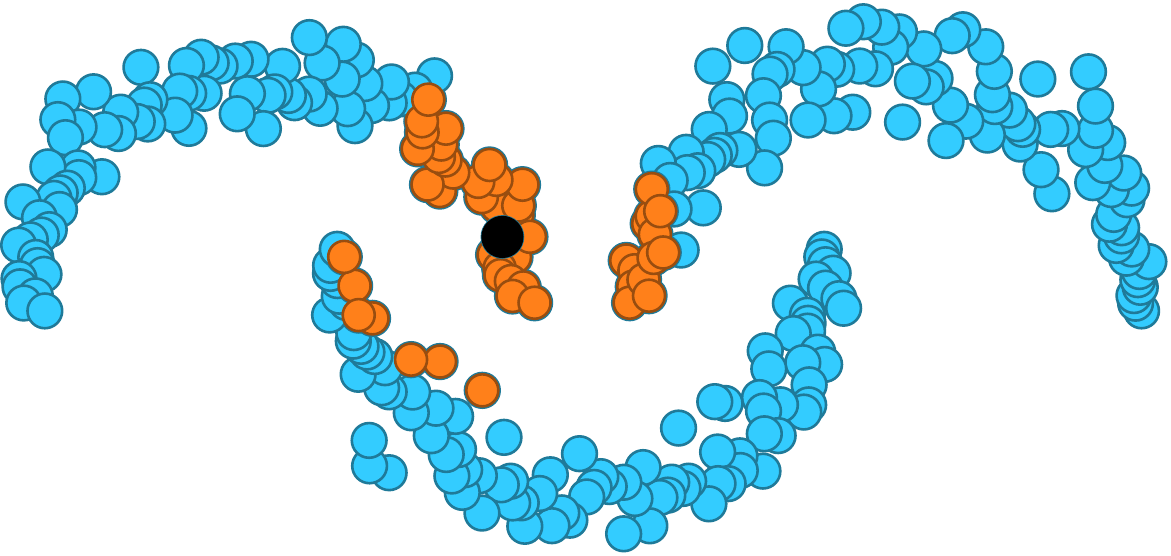}\hspace{6pt}
\includegraphics[height=1.7cm]{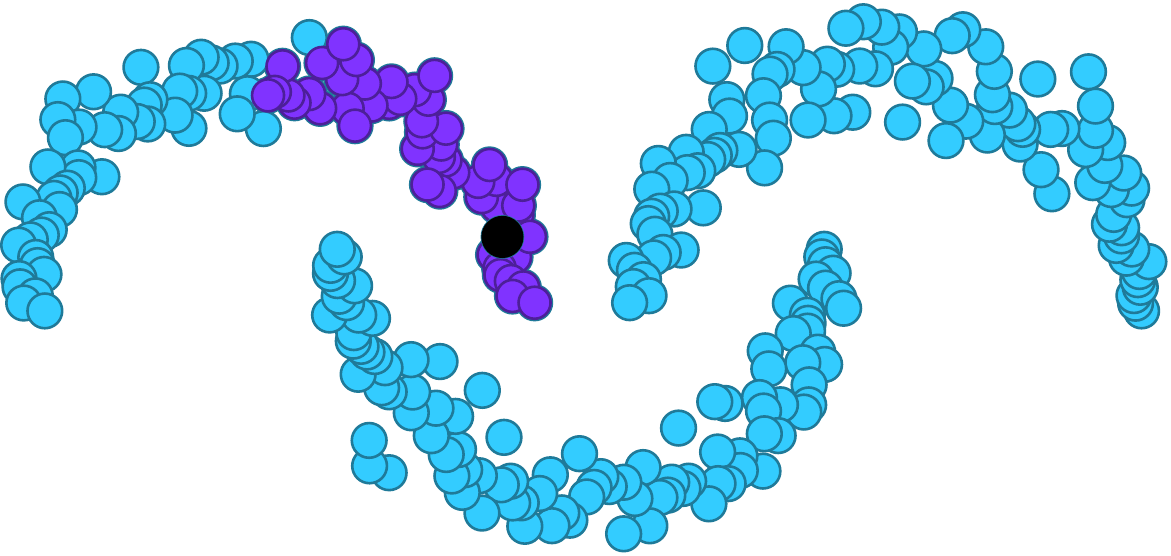}\\
\hspace{5pt}(a) Euclidean NN (orange) \hspace{20pt} (b) Manifold NN (purple)\\ \vspace{3pt}
\includegraphics[height=1.7cm]{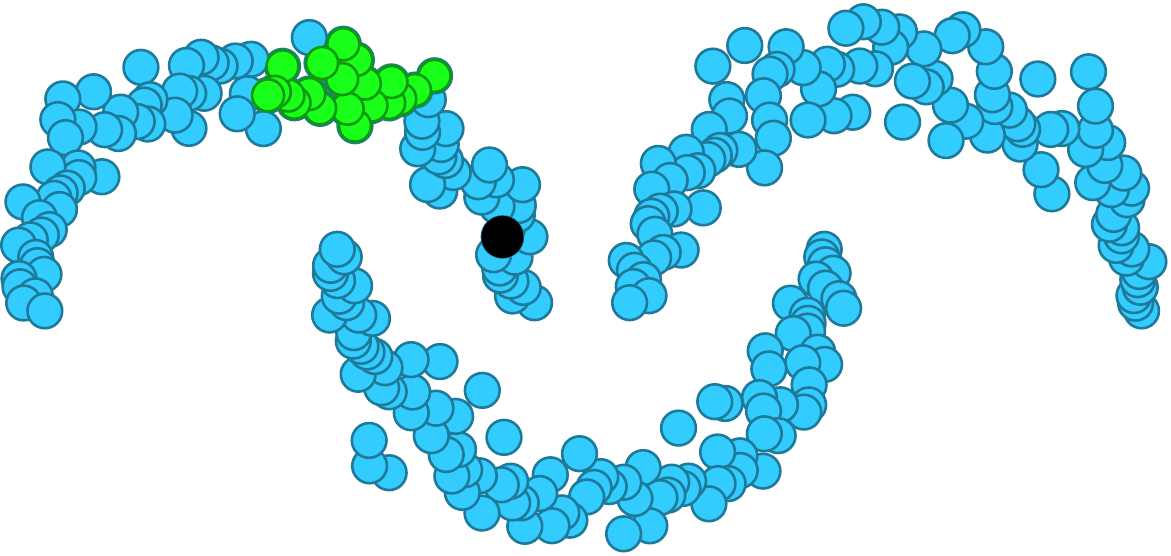}\hspace{6pt}
\includegraphics[height=1.7cm]{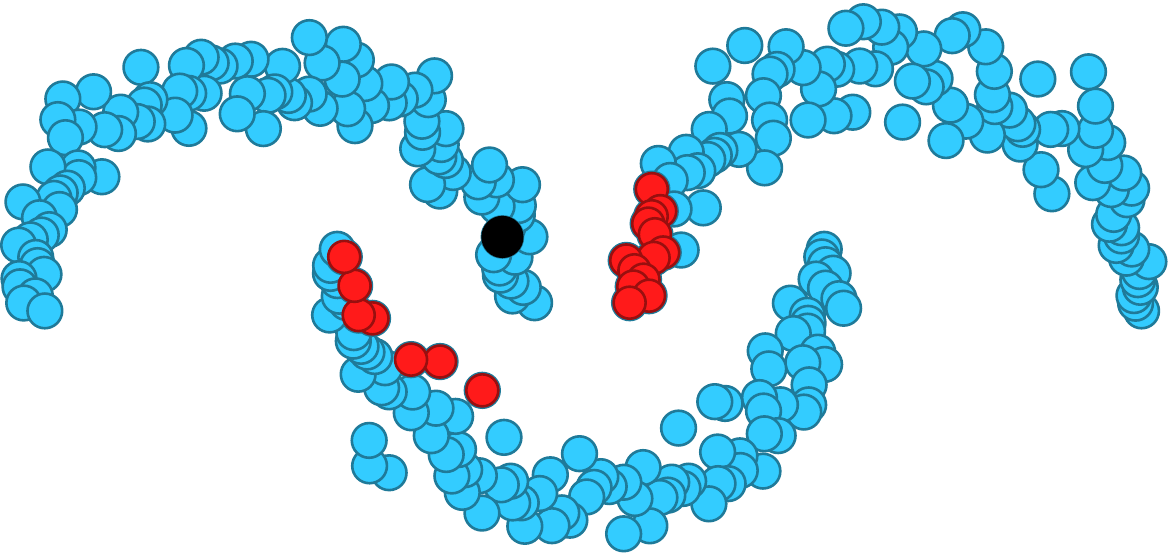}\\
\hspace{5pt}(c) Hard positives (green) \hspace{20pt} (d) Hard negatives (red)
\end{center}
\vspace{-7pt}
\caption{Given an anchor point (black) and its $k$ nearest Euclidean (\nne{k}) and manifold (\nnd{k}) neighbors in a dataset, we choose positive samples as $\nnd{k} \setminus \nne{k}$, and negative as $\nne{k} \setminus \nnd{k}$. Data is unlabeled and the selection is fully unsupervised, including anchors.
\vspace{-15pt}
\label{fig:intro}}
\end{figure}

Most recent deep metric learning approaches can learn powerful embeddings but still use class labels. This is unsatisfying not only because we miss the opportunity of learning from unlabeled data, but learned representations of each class are unimodal~\cite{RPDB15}. Therefore, whatever the loss function~\cite{WSL+14,OXJS16}) or sampling~\cite{HKC+17,MTL+17}, the problem remains supervised classification. On the other hand, conventional nonlinear dimension reduction or manifold learning methods exploit the manifold structure of the data starting from nearest neighbor graphs and are otherwise unsupervised~\cite{TSL00,SR03,BN03}. However, most do not learn an explicit mapping from the input to the embedding space and have difficulties in generalizing to new data.

We attempt to bridge this gap in this work. In particular, we propose a novel method of hard training sample mining in a fully unsupervised manner, simply from an unordered collection of images \emph{relevant} to the final task. We observe that a similarity between two images is improved by considering all, even unlabelled, available data. In particular, we exploit similarity measured on a manifold estimated by a random walk process~\cite{ITA+16}.

The learning starts from the initial representation space of unlabeled data. Given an anchor point that is part of the data, neighbors on the manifold that are not Euclidean neighbors are considered positive samples. In the new learned representation space, the positive sample should be attracted to the anchor to reflect the similarity measured on the manifold. Conversely, Euclidean neighbors that are not neighbors on the manifold are considered negative and should be repelled. The idea is illustrated in Figure~\ref{fig:intro}.

The advantage of learning a new representation over using the estimated manifold similarity is twofold. First, estimating the manifold similarity has additional computational and memory requirements at query time. Second, we show that the novel embedding generalizes better not only to previously unseen queries, but also to unseen datasets.

We apply the proposed method to fine-grained classification and to particular object retrieval. Our models obtained by unsupervised fine-tuning of pre-trained networks are on par or are outperforming prior models that are human supervised or use additional domain-specific expertise.

The paper is organized as follows. Section~\ref{sec:related} discusses related work. Sections~\ref{sec:method} and~\ref{sec:applications} present our learning method and applications, respectively. Experiments are given in Section~\ref{sec:experiments} conclusions are drawn in Section~\ref{sec:conclusions}.

\section{Related work}
\label{sec:related}
This section contains a brief overview of related work on metric learning, embeddings for instance retrieval and representation learning without human labeled data.

\head{Metric learning.}
Recent approaches based on low-dimen\-sional CNN-based embeddings achieve promising results on this task~\cite{SKP15,OXJS16}.
A key ingredient of some approaches is \emph{hard negative mining}, which comes from the days of SVMs in object detection~\cite{FGMR10}. This principle has been known much earlier as \emph{bootstrapping}~\cite{S96}. Instead of sampling all negative instances for an anchor point, the most challenging negative instances are mined which finally offer faster but equally accurate learning.
However, this process is not trivial. Simple hard negative mining from a different class label might corrupt the training process due to mislabeled images~\cite{SKP15,MMRM17}.

Schroff \etal~\cite{SKP15} use triplet loss and propose \textit{semi-hard} mining in an attempt to solve this problem.
They sample negatives within the batch, such that they are close to the anchor point but further away from positives.
This concept is widely adapted in other metric learning approaches~\cite{PVZ15,OXJS16}.
Wu \etal~\cite{WMSK17} improve it by uniformly sampling negative instances weighted by their distance.
This allows them to sample points from various regions of the feature space instead of certain clusters.

Other methods, such as \emph{lifted-structure}~\cite{OXJS16} and $n$-pair loss~\cite{S16}, focus on the loss function and define constraints on the pairwise distances of all points in a batch.
More recently, several methods try to optimize the learned embedding based on the global distribution of the data.
Song \etal~\cite{SJRM17} try to optimize the clustering quality.
Harwood \etal~\cite{HKC+17} combine triplet loss and global loss using an efficient hard mining procedure.
All these methods use class labels during sampling.

\head{Particular object retrieval} is another application where descriptors are trained similarly to metric learning.
However, class labels do not usually exist as it is intractable to enumerate all possible instances or their viewpoints. Traditional instance search algorithms use hand-designed descriptors~\cite{SZ03,PD07,TAJ13,JPDSPS12}, but recent advances show the interest of feature learning~\cite{RTC16,GARL16}. Transfer learning from category-level classification is one case~\cite{RSAC14}. Labeling at landmark level has been attempted as well, treating this task as classification, but the labels are quite noisy~\cite{BSCL14}.

The state-of-the-art approaches start with an off-the-shelf network, and fine-tune it with algorithmic supervision~\cite{RTC16,GARL16}. They make use of geometric matching to mine matching and non-matching image pairs.
They involve local feature extraction and require an expensive pre-processing of data.
Furthermore, they assume that the training set contains landmarks and buildings that perform well with geometric matching of local features. We make no such assumptions nor require additional computer vision system to perform the mining. Our only assumption is that there are multiple object instances in the training set.

\head{Incomplete, noisy, or unavailable labels}
are handled in various ways in the literature.
In the contrastive loss paper of Hadsell \etal~\cite{HaCL06}, the authors show that it is possible to separate different categories to different subspaces just by assigning the Euclidean nearest neighbors  as positive training instances.
In recent works, the labeling is guided by the information of different modalities or the data collection process.
Arandjelovi{\'c} \etal~\cite{AGT+15} learn visual descriptors for location recognition by assigning labels based on the GPS location of each image.
Wang and Gupta~\cite{WG15} sample frames from videos and assume that frames from same videos will be positive to each other.
Isola \etal~\cite{IZKA15} group objects based on their co-occurence within the same spatial or temporal context.
Similarly, learning from the spatial arrangement of the patches within an image is shown feasible~\cite{DGE15,NF16}.
Finally, other cases utilize different modalities, such as learning visual descriptors from text information~\cite{GPRK+17}, or learning audio descriptors from unlabeled videos~\cite{AVT16}.
By contrast, we make no assumptions on the available modalities or contextual information.

\head{Unsupervised methods and manifold learning.}
There are very few methods on deep metric learning that are unsupervised. Examples are two methods on learning fine-grained similarity by exploiting mutual proximity~\cite{BSSO16} and ranking~\cite{BaSO17} in the Euclidean space. Both utilize some form of clustering and splitting the training set in different groups, which is an artificial constraint, and none takes the underlying manifold structure into account. Our method is conceptually simpler and compatible with any loss function requiring positive/negative samples.

The work of Li \etal~\cite{LHH+16} is similar to ours in following a graph-based mining approach. In our comparisons, we show that choosing hard examples is essential and results in better performance for our approach. Recently, Pai \etal~\cite{1711.06011} learn the manifold structure by encouraging pairs of points to have a Euclidean distance in the embedding space equal to the geodesic distance on the graph. We use a more relaxed objective that is compatible to most metric learning formulations and loss functions, and a manifold similarity that is more scalable than the geodesic distance.

Embeddings are learned to approximate ranking on manifolds with \emph{fast spectral ranking}~\cite{ITA+17}. However, this approach is dataset specific and does not generalize to unseen images. Improvements on this aspect are introduced by \emph{iterative manifold embedding}~\cite{JCC+17}. Nevertheless, the extension can handle a query image, or potentially a small set of unseen images, while a larger dataset increase significantly affects the manifold structure.

\section{Method}
\label{sec:method}
In this section we describe our learning problem, briefly discuss the required background, and present our unsupervised training subset selection and the training process.

\subsection{Problem formulation}
Let $X = \{\vx_1, \dots, \vx_n\} \subset \cX$ be an unordered and unlabeled collection of items, where $\cX$ is the original \emph{input} space of all possible items. Depending on the application, an item corresponds to an image, video, image region, local patch, \etc. Function $f(\cdot;\theta): \cX \rightarrow \real^d$ maps an item $\vx$ to a vector $\vz = f(\vx,\theta)$ in a $d$-dimensional \emph{embedding} space, where $\theta$ is a set of parameters to be learned.

The input items are represented by a set of features $Y = \{\vy_1, \dots, \vy_n\} \subset \cY$, where $\cY$ is a \emph{feature} space and $\vy = g(\vx)$ for $\vx \in X$. Depending on the application, $g$ may be the identity mapping, \ie learn directly from input space, $f(\cdot;\theta_0)$, \ie learn from the same model with initial parameters $\theta_0$, or a different model. An existing model may have been supervised (in any way) or not.

Two items are \emph{matching} if they are considered visually similar, otherwise \emph{non-matching}. Our goal is to learn the model parameters such that matching items are mapped to nearby points in the embedding space, while non-matching items are well separated. This corresponds to a typical metric learning scenario, and common practice is to use manually defined labels in order to construct a training set of matching and non-matching pairs of items~\cite{CHL05,WSL+14,HKC+17}. In this work, we only assume that the input items $X$ and their features $Y$ are available at training time.

A training pair is defined \wrt to a \emph{reference} (\emph{anchor}) item $\vx^r$. A matching pair consists of the anchor and a \emph{positive} item $\vx^+$. Similarly a non-matching pair consists of the anchor and a \emph{negative} item $\vx^-$. Alternatively, we may use a triplet $(\vx^r,\vx^+,\vx^-)$. Our goal is to mine such training pairs without any supervision~\cite{CHL05,WSL+14,HKC+17,HKC+17,MTL+17}, without any complementary computer vision system~\cite{GARL16,RTC16} or assumptions on the nature of the training data~\cite{WG15,1708.02901,Doersch_2017_ICCV}.

\subsection{Preliminaries}
By $\nne{k}(\vy)$ we denote the $k$ \emph{Euclidean nearest neighbors} of $\vy \in Y$, \ie the $k$ most similar items in $Y$ according to some \emph{Euclidean similarity} function $\se: \cY^2\to\real$. Similarly, by $\nnd{k}(\vy)$ we denote the $k$ \emph{manifold nearest neighbors} of $\vy \in Y$, \ie the $k$ most similar features in $Y$ according to a \emph{manifold similarity} $\sm:Y^2\to\real$.\footnote{In fact, $\se$ can be any symmetric function but is only a function of two elements of $\cY$; while $\sm$ is a function of two elements in $Y$ but also of the entire set $Y$.}

Given $\se$, we employ a random walk on the Euclidean nearest neighbor graph $G$ to measure the manifold similarity $\sm$~\cite{ZWG+03}. The graph has $Y$ as nodes. It is is undirected, weighted, represented by sparse symmetric adjacency matrix $A = (a_{ij}) \in \real^{n \times n}$. Edges correspond to reciprocal $k$-nearest neighbors, with weights
\begin{equation}
	a_{ij} =
	\begin{cases}
		\se(\vy_i, \vy_j), & \text{if } \vy_i \in \nne{k}(\vy_j) \wedge \vy_j \in \nne{k}(\vy_i) \\
		0, & \text{otherwise.}
	\end{cases}
\label{equ:affinity}
\end{equation}
There are no loops in the graph, \ie, the diagonal elements of $A$ are zero. Starting from an arbitrary vector $\vf^{(0)} \in \real^n$, the random walk for a given feature $\vy_i \in Y$ follows the iteration
\begin{equation}
\vf^{(t)}_i = \alpha \hat{A} \vf^{(t-1)} + (1-\alpha)\ve_i,
\label{equ:walk}
\end{equation}
where $\ve_i$ is the $i$-th column of an $n \times n$ identity matrix, $\hat{A} = D^{-1/2} A D^{-1/2}$ with $D = \diag(A\vone)$ being the degree matrix, and $\alpha \in [0,1)$. Iteration~\eqref{equ:walk} converges to the solution $\vf^{\star}_i$ of the linear system
\begin{equation}
(I-\alpha \hat{A}) \vf  = (1-\alpha)\ve_i.
\label{equ:system}
\end{equation}
Following Iscen \etal~\cite{ITA+16}, we use the conjugate gradient method to solve this system efficiently in practice, since $I-\alpha \hat{A}$ is positive-definite. We define the \emph{manifold similarity}
\begin{equation}
\sm(\vy_i, \vy_j) = \vf^{\star}_i(j),
\label{equ:manifold}
\end{equation}
\ie, the $j$-th element of $\vf^{\star}_i$. Observe that $\sm$ is symmetric because in fact $\sm(\vy_i, \vy_j)$ is the $(i,j)$-element of matrix $(1-\alpha)(I-\alpha \hat{A})^{-1}$, which is symmetric. This matrix is dense but we never compute it; we only compute its columns as needed. For instance, given $\vy_i \in Y$, its manifold nearest neighbors $\nnd{k}(\vy_i)$ are the $k$ maximum elements of the $i$-th column.

\subsection{Manifold-guided selection of training samples}
We are guided by the manifold similarity to select training samples. In particular, we exploit the differences between Euclidean and manifold nearest neighbors of anchor items. We first describe the selection of positives and negatives given an anchor. Then we discuss anchor selection.

\textbf{Positives.} Given an anchor item $\vx^r$ and the corresponding feature $\vy^r = g(\vx^r)$, which is used as query, we choose as positives the items that correspond to the manifold nearest neighbors of $\vy^r$ that are not Euclidean neighbors. Such difference provides evidence of a matching item that is not retrieved well in the feature space, as illustrated in Figure~\ref{fig:intro}(c). In the embedding space, positives should be attracted to the anchor so that Euclidean and manifold neighbors agree.

We therefore simply compare the sets $\nnd{k}(\vy^r)$ and $\nne{k}(\vy^r)$ and select the input items that correspond to their set difference
\begin{equation}
P^+(\vx^r) = \{\vx \in X: g(\vx) \in \nnd{k}(\vy^r) \setminus \nne{k}(\vy^r) \}
\label{equ:positive}
\end{equation}
as the \emph{positive pool} of anchor $\vx^r$. The value of $k$ controls the visual diversity of positives, with larger values giving the \emph{harder} examples. 
In practice, we maintain the pool ordered by descending manifold similarity, so that we may truncate by keeping the examples with highest confidence.

Mining hard positive examples, in contrast to negatives, is not common. Apart from positives being fewer than negatives, this is due to the large intraclass variability; it may result in positives that are too hard to learn. It can be achieved in cases with known geometry of the scene such that extreme cases are avoided~\cite{STF+15,RTC16}. In our case, the hardness is controlled by the manifold similarity according to the current model $g$, so drifting into very tough examples is less likely.

\textbf{Negatives.} Similarly, and as illustrated in Figure~\ref{fig:intro}(d), we choose as negatives the items that correspond to the Euclidean nearest neighbors of $\vy^r$ that are not manifold neighbors. Such difference provides evidence of a non-matching item that is too close in the feature space. In the embedding space, positives should be repelled from the anchor. The \emph{negative pool} of anchor $\vx^r$ is defined accordingly as
\begin{equation}
P^-(\vx^r) = \{\vx \in X: g(\vx) \in \nne{k}(\vy^r) \setminus \nnd{k}(\vy^r) \}.
\label{equ:negative}
\end{equation}
It is common practice, and known to be beneficial~\cite{STF+14}, to select hard negative examples. By construction, its size is controlled by $k$. Again, we maintain the pool ordered by descending Euclidean similarity to keep the hardest negative examples.

\textbf{Anchors.} We are interested in anchors that have many relevant images in the collection, which facilitates propagating on the manifold and discovering differences with the Euclidean neighborhood. We are also interested in anchors that are diverse, so that there is as little redundancy during training. Both conditions are satisfied by the modes of the nearest neighbor graph $G$, which we compute as follows.

We first compute the stationary probability distribution $\vpi$ of a random walk~\cite{ChLe12} on $G$. This is achieved by the power iteration method~\cite{LaMe04} yielding the leading left eigenvector of the transition matrix $P = D^{-1}A$, such that $\vpi P = \vpi$. The probability reflects the \emph{importance} of each node in the graph, as expressed by the probability of a random walker visiting it. We find the local maxima of the stationary distribution on $G$ and out of those, we keep a fixed number having the maximum probability. This is defined as the \emph{anchor set} $\mathcal{A}$. This method has been previously used for image graph visualization~\cite{DoJP16}.

\begin{figure*}
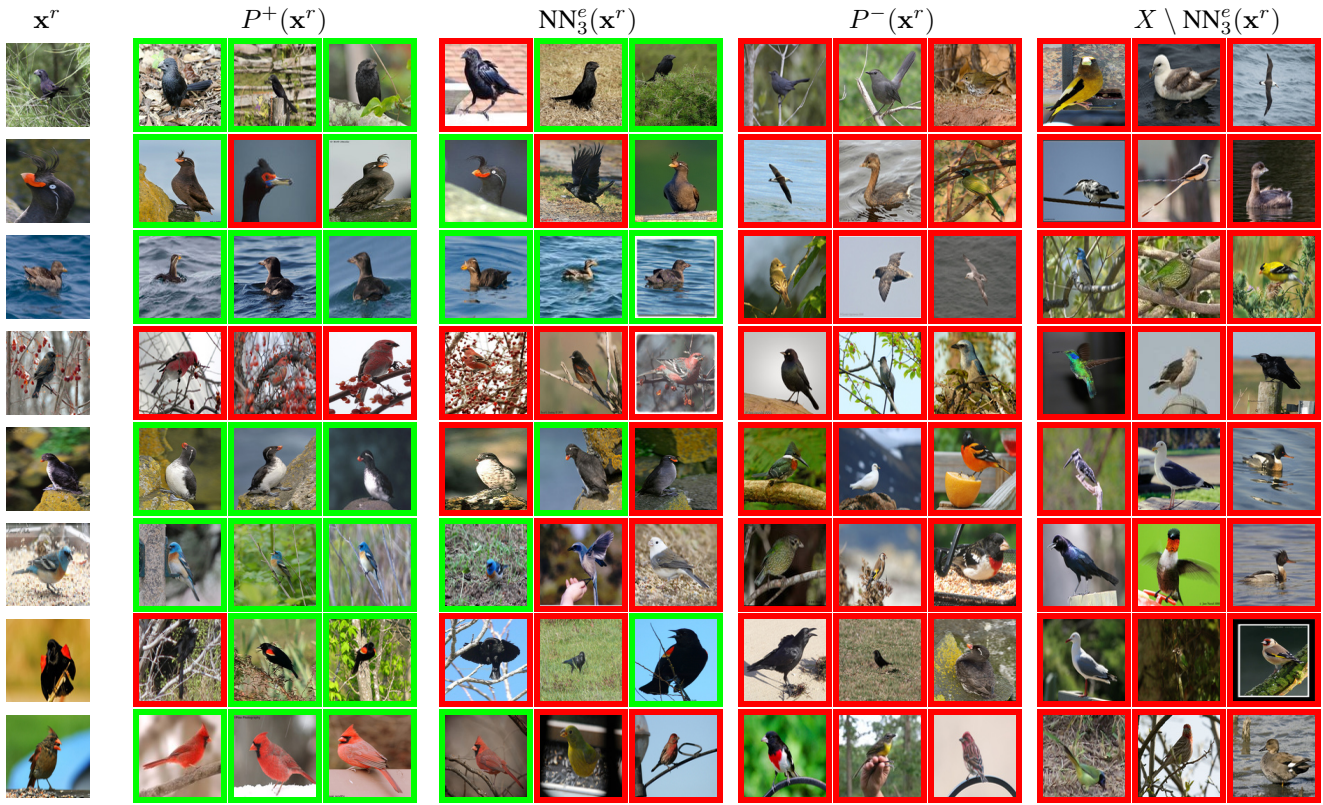

\setlength{\fboxsep}{0pt}%
\setlength{\fboxrule}{2pt}%
\hspace{10pt}$\vx^r$\hspace{68pt}$P^+(\vx^r)$\hspace{80pt}$\nne{3}(\vx^r)$\hspace{80pt}$P^-(\vx^r)$\hspace{75pt}$X\setminus\nne{3}(\vx^r)$\\
 \vspace{-17pt}
\begin{center}
\foreach \q in {231,251,401,571,361,811,501,911}
{
\includegraphics[height=1.1cm, width=1.1cm]{fig/qall_pos50_birds/\q/aq\q} \hspace{5pt}
\pgfplotstableread{fig/qall_pos50_birds/\q/aq\q.txt}\table
\pgfplotstablegetrowsof{\table}
\foreach \i in {1, ..., 3}{
\pgfmathparse{int(\i-1)}\edef\myi{\pgfmathresult}
\pgfplotstablegetelem{\pgfmathresult}{[index]0}\of\table
\ifnum\pgfplotsretval=1
\fcolorbox{green}{black}{\includegraphics[height=1.1cm, width=1.1cm]{fig/qall_pos50_birds/\q/q\q_pos\i}}%
\else
\fcolorbox{red}{black}{\includegraphics[height=1.1cm, width=1.1cm]{fig/qall_pos50_birds/\q/q\q_pos\i}}%
\fi
\hspace{-4pt}}
\hspace{3pt}%
\pgfplotstableread{fig/qall_pos-5_birds/\q/aq\q.txt}\table
\pgfplotstablegetrowsof{\table}
\foreach \i in {2, ..., 4}{
\pgfmathparse{int(\i-1)}\edef\myi{\pgfmathresult}
\pgfplotstablegetelem{\pgfmathresult}{[index]0}\of\table
\ifnum\pgfplotsretval=1
\fcolorbox{green}{black}{\includegraphics[height=1.1cm, width=1.1cm]{fig/qall_pos-5_birds/\q/q\q_pos-\i}}%
\else
\fcolorbox{red}{black}{\includegraphics[height=1.1cm, width=1.1cm]{fig/qall_pos-5_birds/\q/q\q_pos-\i}}%
\fi
\hspace{-4pt}}\hspace{3pt}%
\pgfplotstableread{fig/qall_neg100_birds/\q/aq\q.txt}\table
\pgfplotstablegetrowsof{\table}
\foreach \i in {1, ..., 3}{
\pgfmathparse{int(\i-1)}\edef\myi{\pgfmathresult}
\pgfplotstablegetelem{\pgfmathresult}{[index]0}\of\table
\ifnum\pgfplotsretval=1
\fcolorbox{green}{black}{\includegraphics[height=1.1cm, width=1.1cm]{fig/qall_neg100_birds/\q/q\q_neg\i}}%
\else
\fcolorbox{red}{black}{\includegraphics[height=1.1cm, width=1.1cm]{fig/qall_neg100_birds/\q/q\q_neg\i}}%
\fi
\hspace{-4pt}}\hspace{3pt}%
\pgfplotstableread{fig/qall_neg-5_birds/\q/aq\q.txt}\table
\pgfplotstablegetrowsof{\table}
\foreach \i in {1, ..., 3}{
\pgfmathparse{int(\i-1)}\edef\myi{\pgfmathresult}
\pgfplotstablegetelem{\pgfmathresult}{[index]0}\of\table
\ifnum\pgfplotsretval=1
\fcolorbox{green}{black}{\includegraphics[height=1.1cm, width=1.1cm]{fig/qall_neg-5_birds/\q/q\q_neg-\i}}%
\else
\fcolorbox{red}{black}{\includegraphics[height=1.1cm, width=1.1cm]{fig/qall_neg-5_birds/\q/q\q_neg-\i}}%
\fi\hspace{-4pt}}\\
}
\end{center}
\caption{Sample CUB200-2011 anchor images ($\vx^r$), positive images from our method ($P^+(\vx^r)$) and baseline ($\nne{3}(\vx^r)$), and negative images from our method ($P^-(\vx^r)$) and baseline  ($X\setminus\nne{3}(\vx^r)$). The baseline is Euclidean nearest neighbors and non-neighbors~\cite{HaCL06}.
Positive (negative) ground-truth framed in green (red). Labels are only used for qualitative evaluation and not during training.\label{fig:poolbirds}}
\end{figure*}

\subsection{Training}
The \emph{complete training pool} $\cP$ is the union of the anchor set $\mathcal{A}$ and the positive and negative pools $P^+(\vx), P^-(\vx)$ for each $\vx \in A$. We follow the common practice in metric learning and train a network with two or three branches and use contrastive or triplet loss, respectively. All branches share weights, while the particular network architecture is application specific and is discussed in Section~\ref{sec:applications}.

In both cases of contrastive or triplet loss, we form training tuples of one anchor $\vx^r \in A$, one positive $\vx^+ \in P^+(\vx^r)$, and one negative item $\vx^- \in P^-(\vx^r)$ . At each epoch, the embedding $\vz = f(\vx;\theta)$ for each $\vx \in \cP$, where $\theta$ is the current set of parameters. For each anchor $\vx^r$, a positive item $\vx^+$ is drawn at random from its positive pool, and one negative $\vx^-$ is drawn at random from a subset of its negative pool. This subset consists of the items corresponding to the Euclidean nearest neighbors of $\vz^r = f(\vx^r;\theta)$ in the embedding space. Thus, while the manifold neighbors and the training pool are computed once at the beginning, hard negative sampling uses the current network representation. Finally, the training set for this epoch is the set of such tuples $(\vx^r,\vx^+,\vx^-)$.

Given a tuple $(\vx^r,\vx^+,\vx^-)$, we compute the embeddings $\vz^r=f(\vx^r;\theta)$ $\vz^+=f(\vx^r;\theta)$ and $\vz^-=f(\vx^-;\theta)$, and use the \emph{constrastive loss}~\cite{HaCL06}, combining one positive and one negative pair in a single input,
\begin{equation}
l_c(\vz^r,\vz^+,\vz^-) = \|\vz^r-\vz^+\|^2 + [m - \|\vz^r-\vz^-\|]_+^2,
\label{equ:constrast}
\end{equation}
or the \emph{triplet loss}~\cite{WSL+14}
\begin{equation}
l_t(\vz^r,\vz^+,\vz^-) = [m + \|\vz^r-\vz^+\|^2 - \|\vz^r-\vz^-\|]_+^2,
\label{equ:triplet}
\end{equation}
where $[\cdot]_+$ denotes the positive part and $m$ is a \emph{margin} parameter. We also consider a \emph{weighted} variant of both loss functions, where the loss is multiplied by the manifold similarity $\sm(\vx^r, \vx^+)$ of the positive sample to the anchor. Thus we down-weigh the contribution of the tuples where the positive sample is too hard.

Of course, given the positive and negative pools, there are many more possibilities in sampling positives and negatives and forming losses that are functions of more than two or three items~\cite{LaTC13,OXJS16,BaSO17,KuCR15,UsLe16,WMSK17}. It is also possible to iterate our approach, updating the graph and the pools based on the current embedding space, updating the embeddings, and so on. In this case, given current parameters $\theta$, the set $Z = \{f(\vx;\theta): \vx \in X\}$ plays the role of $Y$ for the following iteration. This alternating training scheme is common for methods involving a global dataset structure like a graph or a set of clusters~\cite{RPDB15,BACH15,JCC+17}. Our idea is orthogonal to most concurrent improvements on metric learning.

\section{Applications}
\label{sec:applications}
We apply the proposed method to learn visual representations on two different tasks: fine-grained categorization~\cite{WCMD16,HKC+17} and instance-based image retrieval~\cite{GARL16,RTC16}. In both cases, both the features $Y$ and the initial model $f(\cdot,\theta)$ are based on a pre-trained model. We assume there are no labeled images available.

\subsection{Fine-grained categorization}
\label{sec:toPoulo}
We use the CUB200-2011 dataset~\cite{WBWP+11} comprising $200$ bird species. The goal is to learn embeddings that better discriminate instances of the same class from instances of different classes. Following the setup of~\cite{OXJS16}, the training set contains half ($100$) classes on which embedding is learned, while the rest are used for testing. Given a test query image, the remaining test images of the same species should be top-ranked \wrt Euclidean distance to the query.

All prior approaches are fully supervised and use the manually assigned
labels of the training set. Our method is unsupervised, but otherwise we choose the same settings with the literature in our comparisons. In particular, we initialize the network by GoogLeNet~\cite{SLJS+15} as pre-trained on ImageNet and add a fully connected layer right after the average pooling layer, reducing the embedding dimensionality to $d=64$. We perform training with the triplet loss using all training images as anchors, which
is affordable due to the small size ($6$k images) of the training set.

Our initial features are formed by R-MAC~\cite{TSJ15} on the last convolutional feature map of the pre-trained GoogLetNet, right before the average pooling layer, aggregated over 3 input image scales and whitened.
The feature set $Y$ contains all such vectors $\vy = g(\vx)$ for $\vx$ in the entire training set $X$.
In Figure~\ref{fig:poolbirds} we show examples of anchors and subsets of their positive and negative pools. Despite the absence of labels and the challenges of fine-grained similarity, we achieve a very clean negative pool and a reasonably clean positive one.

\pgfmathsetseed{103}
\begin{figure*}
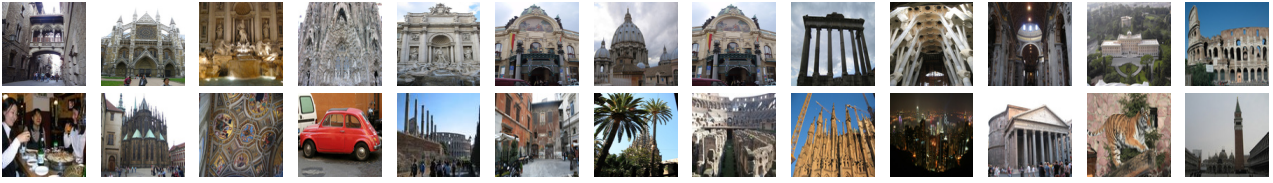

\begin{center}
\foreach \m in {85,1,99,56,7,17,19,17,38,64,83,75,18}{ \includegraphics[height=1.1cm, width=1.1cm]{fig/pagerank_seeds/query\m} \hspace{1pt}}\\ \vspace{2pt}
\foreach \m in {603,229,205,424,684,679,910,131,251,334,65,670,647}{ \includegraphics[height=1.1cm, width=1.1cm]{fig/pagerank_seeds/query\m} \hspace{1pt}}\\
\end{center}
\vspace{-6pt}
\caption{Examples of anchor images selected by the proposed method for the image retrieval application. Random samples from the top 100 (1000) anchors are shown in the top (bottom) row according to the node importance.\label{fig:anchors}}
\end{figure*}

\begin{figure*}
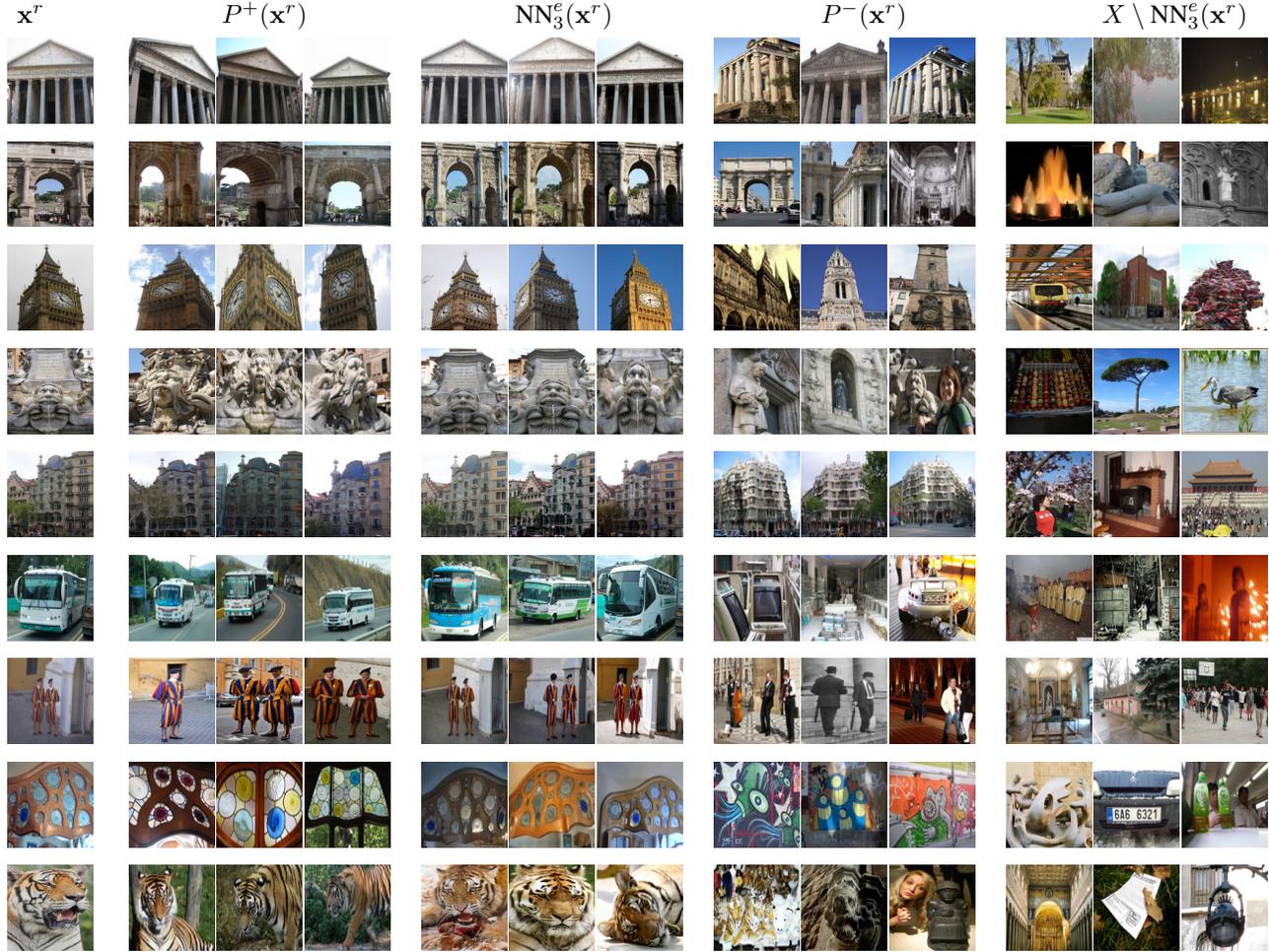

\setlength{\fboxsep}{1pt}%
\setlength{\fboxrule}{1pt}%
\hspace{10pt}$\vx^r$\hspace{68pt}$P^+(\vx^r)$\hspace{80pt}$\nne{3}(\vx^r)$\hspace{80pt}$P^-(\vx^r)$\hspace{75pt}$X\setminus\nne{3}(\vx^r)$\\
 \vspace{-17pt}
 \begin{center}
\foreach \q in {58,81,107,135,225,465,541,573,581}
{
\fcolorbox{white}{white}{\includegraphics[height=1.15cm, width=1.15cm]{fig/pagerank_pos50/\q/aq\q}}\hspace{10pt}%
\fcolorbox{white}{white}{\foreach \i in {1, ..., 3}{\includegraphics[height=1.15cm, width=1.15cm]{fig/pagerank_pos50/\q/q\q_pos\i}\hspace{1pt}}}\hspace{7pt}%
\fcolorbox{white}{white}{\foreach \i in {2, ..., 4}{\includegraphics[height=1.15cm, width=1.15cm]{fig/pagerank_pos-5/\q/q\q_pos-\i}\hspace{1pt}}}\hspace{7pt}%
\fcolorbox{white}{white}{\foreach \i in {1, ..., 3}{\includegraphics[height=1.15cm, width=1.15cm]{fig/pagerank_neg10000/\q/q\q_neg\i}\hspace{1pt}}}\hspace{7pt}%
\fcolorbox{white}{white}{\foreach \i in {1, ..., 3}{\includegraphics[height=1.15cm, width=1.15cm]{fig/pagerank_neg-5/\q/q\q_neg-\i}\hspace{1pt}}}\\\vspace{2pt}%
}
\end{center}
\vspace{-6pt}
\caption{Sample anchor images ($\vx^r$), positive images from our method ($P^+(\vx^r)$) and baseline ($\nne{3}(\vx^r)$), and negative images from our method ($P^-(\vx^r)$) and baseline  ($X\setminus\nne{3}(\vx^r)$). The baseline is Euclidean nearest neighbors and non-neighbors~\cite{HaCL06}.
\label{fig:poolretrieval}}
\end{figure*}

\subsection{Particular object retrieval}
\label{sec:retrieval}
Particular object retrieval differs from bird species classification
in that
images
are instances of the same object and not of the same (sub-)category, \ie the similarity is even more fine-grained than in bird species. Objects are less deformable, but there is extreme diversity in viewpoint, illumination conditions, occlusion and background clutter.

Methods trained with category-level labeling do not perform well~\cite{BSCL14}, while the state-of-the-art approaches use direct geometric matching~\cite{GARL16} or Structure-from-Motion (SfM)~\cite{RTC16} to automatically mine matching and non-matching image pairs. This is appropriate since the geometry of the scene is known, but the whole process assumes the existence of another computer vision system based on local descriptors, which is rather expensive~\cite{SRCF15}.

To be comparable to the state-of-the-art fine-tuned MAC obtained through SfM by Radenovic \etal~\cite{RTC16}, we use similar training set and design choices. In particular, we start from the same set of 7M images, referred to as \emph{Flickr 7M} in the following\footnote{Images depicting buildings that are part of the Oxford5k or Paris6k test set are removed as in~\cite{RTC16}.}, which is downloaded from Flickr with text tag queries. We limit the training set by randomly sampling 1M images. We initialize our network by VGG~\cite{SZ14} as pre-trained on ImageNet and we fine-tune MAC representation with contrastive loss.

Our initial features are formed by R-MAC~\cite{TSJ15} on the last convolutional feature map of the pre-trained VGG network. The feature set $Y$ contains all such vectors $\vy = g(\vx)$ for $\vx$ in the entire training set $X$. We sample an anchor set $\mathcal{A}$ and construct the complete training pool $\cP$. Anchor selection is essential here, as using all images as queries would be very expensive. Compared to~\cite{RTC16}, our complete training pool is larger (50k vs 22k), however, we only choose 1k anchors per epoch vs 6k.

Examples of selected anchors are shown in Figure~\ref{fig:anchors}. They usually correspond to popular locations frequently appearing in the dataset.
Examples of training pools are shown in Figure~\ref{fig:poolretrieval}, comparing to a baseline where positives and negatives are Euclidean nearest neighbors and non-neighbors, respectively.
The same baseline is used for comparisons in our experiments. The baseline positives contain only mild viewpoint and illumination changes, while negatives are random. On the contrary, our positives contain more challenging changes and very interesting negatives: different objects, which still look similar in one way or another.
The training set variability, in terms of different objects and viewing conditions, seems a desirable property.

\subsection{Implementation details}
We always create graph $G$ by considering $k=30$ nearest neighbors in (\ref{equ:affinity}). Exact computation takes $80$min on $12$ CPU threads on the 1M retrieval training set. We use the Euclidean similarity function $\se(\vx_i, \vx_j)=[\vx_i^\top \vx_j]_+^3$ and $\alpha=0.99$ following Iscen~\etal~\cite{ITA+16}. In order to create the positive pool we consider 50 neighbors in (\ref{equ:positive}), while for the negative pool we use 100 and 10,000 neighbors in (\ref{equ:negative}) for fine-grained categories and retrieval, respectively, due to the different size of the training dataset. We finally restrict the negative pool of each anchor to have 50 instances at most. All vector representations used for an image in the feature and the embedding space are $\l2$-normalized.

We use stochastic gradient descent with momentum for optimization. The learning rate is initialized at $10^{-2}$, and scaled by $0.1$ every $10$ epochs. The momentum parameter is $0.9$. The margin $m$ is set to $0.5$ for triplet loss in fine-grained categorization, and to $0.7$ for contrastive loss in particular object retrieval.
The batch size includes $42$ triplets for fine-grained categories~\cite{OXJS16} and pairs for $5$ anchors in the retrieval application~\cite{RTC16}. We train for 100 epochs on fine-grained categorization and 30 epochs on particular object retrieval experiments.

\section{Experiments}
\label{sec:experiments}
Fine-grained categorization is evaluated on CUB200-2011~\cite{WBWP+11}, where we use the training set without labels for training and then evaluate on the test set, measuring Recall@$k$ as well as clustering quality by NMI~\cite{MRS08}. Particular object retrieval is evaluated by mean average precision (mAP) on a challenging and diverse set of test datasets comprising landmark and building images (Oxford5k~\cite{PCISZ07} and Paris6k~\cite{PCISZ08}), natural landscapes (Holidays~\cite{JDS08}), as well as planar and 3D objects (Instre~\cite{WJ15}). Large scale experiments are performed on Oxford and Paris by adding 100k distractors~\cite{PCISZ07}, namely Oxford105k and Paris106k respectively.
We first evaluate the importance of different components of the selection strategy and then compare our method against state-of-the-art on each task.

\begin{table}
\vspace{5pt}
\begin{center}
\def\w{\hspace{-2pt}+\hspace{-2pt}W\hspace{-2pt}}
\small
\begin{tabular}{ |@{\ssp}l@{\ssp}|@{\ssp}l@{\ssp}|c|@{\ssp}c@{\ssp}|@{\ssp}c@{\ssp}|}
\hline
      Positive            & Negative               & CUB             & \multicolumn{2}{c|}{Oxford5k}      \\ \hline
      \multicolumn{2}{|@{\nsp}c@{\nsp}|}{Anchors}  & All             & Random          & {$\cA$ }         \\ \hline \hline
      \multicolumn{2}{|@{\nsp}c@{\nsp}|}{Initial}  & 35.0            & \multicolumn{2}{c|}{52.6}          \\  \hline
      \nne{5}             & $X \setminus$\nne{5}   & 38.5            & 37.4            & 41.9             \\
      $P^+$               & $X \setminus$\nne{5}   & 43.0            & 48.2            & 38.1             \\
      \nne{5}             & $P^-$                  & 42.1            & 57.8            & 71.3             \\
      $P^+$               & $P^-$                  & 43.5            & 64.4            & 73.7             \\
      $P^+ \w$            & $P^-\w$                & \textbf{45.3}   & 67.0            & \textbf{76.7}    \\ \hline
\end{tabular}
\caption{Impact of choices of anchors and pools of positive and negative examples on Recall@1 on CUB-200-2011 and mAP on Oxford5k. On CUB, all images are used as anchors, while on Oxford5K anchors are selected either at random or by the proposed method ($\cA$).
The positive and negative pools are formed by either the baseline with Euclidean nearest neighbors (\nne{5})~\cite{HaCL06} or our selection ($P^+$ and $P^-$), optionally with our weighted loss ($\w$).
\vspace{-15pt}
\label{tab:allSel}}
\end{center}
\end{table}

\begin{table}
\begin{center}
\small
\begin{tabular}{ |@{\sssp}l@{\sssp}|@{\sssp}c@{\sssp}|@{\sssp}c@{\sssp}|@{\sssp}c@{\sssp}|@{\sssp}c@{\sssp}|@{\sssp}c@{\sssp}|@{\sssp}c@{\sssp}|}
	\hline
	Method                               & Labels   & R@1     & R@2    & R@4    & R@8    & NMI    \\ \hline  \hline
	Initial                              & No       & 35.0    & 46.8   & 59.3   & 72.0   & 48.1   \\ \hline
	Triplet+semi-hard~\cite{SKP15}       & Yes      & 42.3    & 55.0   & 66.4   & 77.2   & 55.4   \\
	Lifted-Structure (LS)~\cite{OXJS16}  & Yes      & 43.6    & 56.6   & 68.6   & 79.6   & 56.5   \\
	Triplet+~\cite{HKC+17}               & Yes      & 45.9    & 57.7   & 69.6   & 79.8   & 58.1   \\
	Clustering~\cite{SJRM17}             & Yes      & 48.2    & 61.4   & 71.8   & 81.9   & 59.2   \\
	Triplet+++\cite{HKC+17}              & Yes      & 49.8    & 62.3   & 74.1   & 83.3   & 59.9   \\ \hline
	Cyclic match\cite{LHH+16}            & No       & 40.8    & 52.8   & 65.1   & 76.0   & 52.6   \\
	Ours                                 & No       & 45.3    & 57.8   & 68.6   & 78.4   & 55.0   \\ \hline
\end{tabular}
\caption{Recall@$k$ and NMI on CUB-200-2011. All methods expect for ours and cyclic match~\cite{LHH+16} use ground-truth labels during training.
\label{tab:birdsSoa}}
\vspace{-15pt}
\end{center}
\end{table}

\begin{table*}[t!]
\vspace{1pt}
\begin{center}
\begin{tabular}{ |@{\msp}l@{\msp}|@{\msp}l@{\msp}|@{\msp}l@{\msp}|@{\msp}c@{\msp}|@{\msp}c@{\msp}|@{\msp}c@{\msp}|@{\msp}c@{\msp}|@{\msp}c@{\msp}|@{\msp}c@{\msp}|}
	\hline
	Method                     &	Representation        & Labels        & Oxford5k       & Oxford105k     & Paris6k         & Paris106k      & Holidays        & Instre          \\ \hline  \hline
	Pre-trained~\cite{TSJ15}   & \multirow{3}{*}{MAC}   & ImageNet      & 58.5           & 50.3           & 73.0            & 59.0           & 79.4            & 48.5            \\
	CNN from BoW~\cite{RTC16}  &                        & SfM           & \textbf{79.7}  & 73.9           & 82.4            & 74.6           & 81.4            & 48.5            \\
	Ours                       &                        & No            & 78.7           & \textbf{74.3}  & \textbf{83.1}   & \textbf{75.6}  & \textbf{82.6}   & \textbf{55.5}   \\ \hline
	Pre-trained~\cite{TSJ15}   & \multirow{3}{*}{R-MAC} & ImageNet      & 68.0           & 61.0           & 76.6            & 72.1           & 87.0            & 55.6            \\
	CNN from BoW~\cite{RTC16}  &                        & SfM           & 77.8           & 70.1           & 84.1            & 76.8           & 84.4            & 47.7            \\
	Ours                       &                        & No            & \textbf{78.2}  & \textbf{72.6}  & \textbf{85.1}   & \textbf{78.0}  & \textbf{87.5}   & \textbf{57.7}   \\ \hline
\end{tabular}
\caption{mAP on particular object retrieval datasets. We compare VGG as pre-trained on ImageNet, the fine-tuned network of Radenovic~\etal~\cite{RTC16}, and our fine-tuned one. Fine-tuning is always performed for MAC, but at testing we evaluate both global MAC and regional R-MAC~\cite{TSJ15} representations.\label{tab:retrSoa}}
\vspace{-5pt}
\end{center}
\end{table*}

\subsection{Impact of positives, negatives, and anchors}
We consider the unsupervised approach proposed by Hadsell \etal~\cite{HaCL06} as a baseline method.
It sets positives to be the 5 nearest neighbors with Euclidean distance. All other elements are considered negatives out of which we randomly draw the negative pool of an anchor.
In this case, hard negative mining per epoch is impossible and random choice is the only choice.
We present the results in Table~\ref{tab:allSel}.
Our method improves the pre-trained network in both tasks without any supervision, while the weighted loss consistently helps.
Furthermore, we observe that our hard negatives are beneficial and necessary, while our anchors are essential for the large scale training set.
Given popular anchors, even simple nearest neighbors work well as positives. However, our harder positives further improve.

CUB's annotation allows us to measure the true positive and true negative rate in our positive and negative pools.
These measurements are $40\%$ and $96\%$, respectively.
Additionally, we exploit the labels and train with a \emph{positive (negative) oracle}, where we replace our positive (\emph{resp}. negative) pool with one based on labels, yielding 46.7 (\emph{resp}. 36.5) Recall@1.
This shows that our hard positives with weighting are almost as good as true positives and that our hard negatives are better than hard annotated negatives.
The latter is due to errors in annotation of CUB dataset, quantified to 4.4\%~\cite{VBF+15}. In this case, hard negative mining frequently finds false negative examples, which are known to cause training failure~\cite{SKP15}.

\subsection{Comparisons on fine-grained categories}

To our knowledge there is no other unsupervised method that evaluates on CUB200-2011 dataset.
We evaluate the unsupervised approach of Li \etal~\cite{LHH+16} by constructing the same graph as in our method, then using the provided code to construct the positive/negative pool and finally training the same way as in our method.
We also compare to supervised methods that use ground-truth labels on the training set, but otherwise an identical experimental setup.
As shown in in Table~\ref{tab:birdsSoa}, our method competes or even outperforms fully supervised methods.
We also outperform the only unsupervised competitor. Although their true positive rate is higher (76\%), their positive pairs are mostly extremely similar and not challenging enough for training.
Note that there are better-performing methods in the literature with sophisticated sampling schemes~\cite{WMSK17}, which can be complementary to ours.

\subsection{Comparisons on retrieval}
We initialize by VGG as pre-trained on ImageNet and fine-tune using MAC representation. At testing, the fine-tuned network is evaluated with both global MAC and regional R-MAC representations~\cite{TSJ15}.
This is the same process as~\cite{RTC16}, which makes it comparable in this respect, although the training set and sampling pool sizes are not the same as discussed in section~\ref{sec:retrieval}.

Descriptor whitening is known to be essential for MAC and R-MAC. We follow the common practice in the literature and perform unsupervised PCA whitening~\cite{JC12} for the pre-trained networks, and supervised LDA-based whitening for the fine-tuned networks, learned on a subset of \emph{Flickr 7M}. In particular, as supervision we use SfM labels for the network of Radenovic~\etal~\cite{RTC16}, and matching pairs consisting of \nnd{50} per anchor for our method.

As shown in Table~\ref{tab:retrSoa}, we improve over the pre-trained network on all test sets. Moreover, we outperform~\cite{RTC16} with only one exception on Oxford5k.
Remarkably, we perform better even on building or landmark oriented test sets, while their method specifically favors this kind of images.
With our training pool being more diverse, we improve on Holidays and Instre test sets, where ~\cite{RTC16} shows little improvement or is even inferior to the pre-trained network.

\section{Conclusions}
In this work, we depart from using discrete category level labeling in order to learn fine grained similarity.
Not only we avoid the expensive or nearly impossible manual annotation, but also do not restrict the problem to supervised classification.
Our unsupervised and manifold-aware sampling of training data is applied to perform metric learning. The learning attracts points that lie on the same manifold and repels points on different manifolds. The method is conceptually simple and applicable with standard contrastive and triplet loss. It is shown to be effective for fine-grained categorization and particular object retrieval, competing or surpassing fully supervised approaches.
\label{sec:conclusions}

\head{Acknowledgments}
The authors were supported by the MSMT LL1303 ERC-CZ grant.

{\small
\bibliographystyle{ieee}
\bibliography{egbib}
}
\end{document}